\documentclass{article}
\usepackage{graphicx}

\usepackage[utf8]{inputenc}
\usepackage{amsmath,amsfonts,amssymb,dsfont}
\usepackage{multirow}
\usepackage{afterpage}
\usepackage{pifont} 
\newcommand{\cmark}{\ding{51}}%
\newcommand{\inlineeqnum}{\refstepcounter{equation}~~\mbox{(\theequation)}}

\begin{document}
\title{From Classical to Generalized Zero-Shot Learning: a Simple Adaptation Process}
%
%
\author{
\textbf{Yannick Le Cacheux}\\CEA LIST
\and
\textbf{Hervé Le Borgne}\\CEA LIST
\and
\textbf{Michel Crucianu}\\CEDRIC Lab -- CNAM
}
%
%
%
\maketitle
\begin{abstract}
Zero-shot learning (ZSL) is concerned with the recognition of previously \textit{unseen} classes. 
It relies on additional semantic knowledge for which a mapping can be learned with training examples of \textit{seen} classes. While classical ZSL considers the recognition performance on unseen classes only, generalized zero-shot learning (GZSL) aims at maximizing performance on both seen and unseen classes. In this paper, we propose a new process for training and evaluation in the GZSL setting;
this process addresses the gap in performance between samples from unseen and seen classes by penalizing the latter, and enables to select hyper-parameters well-suited to the GZSL task. It can be applied to any existing ZSL approach and leads to a significant performance boost:
the experimental evaluation shows that GZSL performance, averaged over eight state-of-the-art methods, is improved from $28.5$ to $42.2$ on CUB and from $28.2$ to $57.1$ on AwA2.
\end{abstract}
\section{Introduction}
Zero-shot learning (ZSL)~\cite{lampert2009,larochelle2008,palatucci2009} aims to recognize classes for which no training example is available. This is often achieved by relying on additional semantic knowledge, consisting for example in vectors of attributes.
During training, a relation between visual features and semantic attributes is learned from training examples belonging to the \emph{seen} classes, for which both modalities (visual and semantic) are available. 
This model is then applied in the testing phase on examples from \emph{unseen} classes,
for which no visual instance was available during training.
Predictions on these classes can thus be made on the basis of the inferred relation between visual and semantic features.

In classical ZSL, the test set only contains examples from the novel, unseen classes, and these classes alone can be predicted. Although this setting has enabled significant progress in methods linking visual content to semantic information in the last few years~\cite{xian2017cvpr}, it is hardly realistic. It seems much more reasonable to assume that objects which are to be classified can belong to either a seen class or an unseen class, since in real-life use-cases one could legitimately want to recognize both former and novel classes. This setting is usually referred to as generalized zero-shot learning (GZSL).

However, recent work shows that a direct use of a ZSL model in a GZSL setting usually leads to unsatisfactory results. 
Indeed, in addition to the number of candidate classes being higher due to the presence of the seen classes among them, most samples from unseen classes are incorrectly classified as belonging to one of the seen classes~\cite{chao2016}. 
Different methods have been proposed to measure this discrepancy, such as the area under the curve representing all the possible trade-offs between the accuracies on samples from seen classes versus samples from unseen classes~\cite{chao2016}, or their harmonic mean~\cite{xian2017cvpr} to penalize models with strong imbalance between the two. While these proposals only measure the extent of the problem, we aim to explicitly address this issue in addition to quantifying its impact. 

The main contribution of this paper is a new process for training and evaluating models in a GZSL setting. In accordance with recent studies, we show that the application of a ZSL model ``out of the box'' gives results that are far from optimal in the GZSL context. We demonstrate how two simple techniques -- the calibration of similarities and the use of appropriately balanced regularization -- can dramatically improve the performance of most models. The final score for the GZSL task can thus be increased up to a factor of two, with no change regarding the underlying hypotheses of the GZSL task or the data available at any given time, which means that our process is applicable to any ZSL model. We also provide new insights on the reasons why these two techniques are relevant and on the fundamental differences between samples from seen and unseen classes.

We extensively evaluate these techniques on several recent ZSL methods. For sanity-check, we independently reproduce results obtained in the literature before applying our process. We find that some models show a variability in performance with respect to their random initialization, so measures averaged over several runs should be preferred. We eventually find that, with fair comparison under unbiased conditions as enabled by our process, a regularized linear model can give results close to or even better than the state-of-the-art.

\begin{figure}
\includegraphics[width=\textwidth]{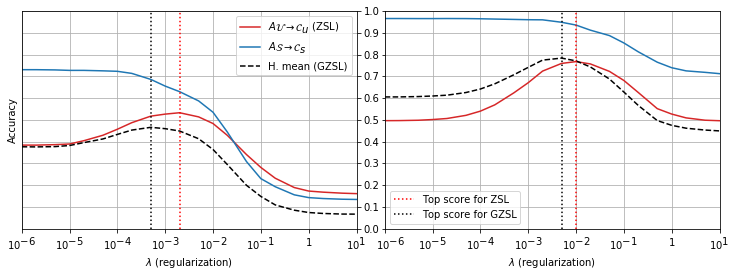}
\caption{Illustration of how the regularization parameter $\lambda$ affects the accuracies on samples from seen and unseen classes $A_{\mathcal{U} \rightarrow \mathcal{C}_u}$ and $A_{\mathcal{S} \rightarrow \mathcal{C}_s}$ (see Sec.~\ref{sec:statement}) as measured on CUB~\cite{cub} (left) and AwA2~\cite{xian2017arxiv} (right). Optimal regularization is not the same in a ZSL setting, where performance is measured by $A_{\mathcal{U} \rightarrow \mathcal{C}_u}$ (red dotted line), and in a GZSL setting, where it is measured by the harmonic mean of $A_{\mathcal{U} \rightarrow \mathcal{C}}$ and $A_{\mathcal{S} \rightarrow \mathcal{C}}$ (black dotted line).}
\label{fig:illustration}
\end{figure}

\section{Related work}

An early rigorous definition and evaluation of GZSL was put forward in~\cite{chao2016}. The authors argue that this setting is more realistic than ZSL and highlight the gap between accuracies on seen and unseen classes when labels from \emph{all} classes can be predicted (denoted respectively $A_{\mathcal{U} \rightarrow \mathcal{C}}$ and $A_{\mathcal{S} \rightarrow \mathcal{C}}$, and formally defined in Sec.~\ref{sec:statement}). They also introduce the idea of calibration to address this issue and suggest a new metric for GZSL, \emph{Accuracy Under Seen-Unseen Curve (AUSUC)}, which measures the trade-off between the two accuracies but does not directly provide the expected performance in real use-cases.

An extensive evaluation of recent ZSL methods with a common protocol is provided in~\cite{xian2017cvpr}, both in ZSL and GZSL settings. The authors use a different metric for GZSL, the harmonic mean between $A_{\mathcal{U} \rightarrow \mathcal{C}}$ and $A_{\mathcal{S} \rightarrow \mathcal{C}}$, which does not directly quantify the trade-off between accuracies but better estimates the practical performance of a given model. 
However, they do not explicitly address the gap between similarities evaluated on seen and unseen classes~\cite{chao2016},
which has a significant impact on the final performance as we show in Sec.~\ref{sec:results}. 

%
Further GZSL results based on the harmonic mean metric are provided in~\cite{bucher2017,verma2018,xian2018}. All three methods rely on generators of artificial training examples from unseen classes. However, these methods assume that a semantic description of all unseen classes is available during training. This assumption is not necessarily met in practice and makes the inclusion of additional unseen classes more difficult.

Transductive ZSL methods~\cite{fu2015,kodirov2015,rohrbach2013} also assume that additional information, taking the form of unlabeled samples from unseen classes, is available during training. This can naturally lead to improved performance. In this article, we make none of these assumptions and consider that \emph{no} information regarding unseen classes is available at training time.

\section{Proposed approach}

\subsection{Problem statement} \label{sec:statement}

We denote by $\mathcal{C}_s$ the set of classes \emph{seen} during training and by $\mathcal{C}_u$ the set of \emph{unseen} classes. We define $\mathcal{C} = \mathcal{C}_s \cup \mathcal{C}_u$, with $\mathcal{C}_s \cap \mathcal{C}_u = \emptyset$.
During the training phase, we consider $N^{tr}$ training samples consisting of $D$-dimensional visual features $\mathbf{X}^{tr} = (\mathbf{x}^{tr}_1, \dots , \mathbf{x}^{tr}_{N^{tr}})^\top \in \mathbb{R}^{N^{tr}\times D}$ and corresponding labels $\mathbf{y}^{tr} = (y^{tr}_1, \dots , y^{tr}_{N^{tr}})^\top \in {\mathcal{C}_s}^{N^{tr}}$, as well as $K$-dimensional semantic \emph{class prototypes} noted by $\mathbf{S}^{tr} = (\mathbf{s}^{tr}_1, \dots , \mathbf{s}^{tr}_{|\mathcal{C}_s|})^\top \in \mathbb{R}^{|\mathcal{C}_s| \times K}$.
We seek to learn a function $f: \mathbb{R}^D \times \mathbb{R}^K \rightarrow \mathbb{R}$ assigning a similarity score to each pair composed of a visual feature vector and a semantic representation so as to minimize the following regularized loss:
\begin{equation} \label{eq:objective}
\frac{1}{N^{tr}} \sum_{n=1}^{N^{tr}} \sum_{c=1}^{|\mathcal{C}_s|} L(f(\mathbf{x}^{tr}_n; \mathbf{s}^{tr}_c), y^{tr}_n) + \lambda\Omega[f]
\end{equation}
where $L$ is the loss function and $\Omega$ the regularization term weighted by $\lambda$.
During the testing phase, we consider $N^{te}$ unlabeled visual samples $\mathbf{X}^{te} = (\mathbf{x}^{te}_1, \dots , \mathbf{x}^{te}_{N^{te}})^\top \in \mathbb{R}^{N^{te}\times D}$ and class prototypes for \emph{candidate classes}. In a ZSL setting, the candidate classes $\mathcal{C}^{te}$ are the unseen classes such that $\mathbf{S}^{te} \in \mathbb{R}^{|\mathcal{C}_u| \times K}$.
In a GZSL setting, classes to be predicted can be in either $\mathcal{C}_u$ or $\mathcal{C}_s$, such that $\mathcal{C}^{te} = \mathcal{C}$ and the class prototypes are $\mathbf{S}^{te} = (\mathbf{s}^{te}_1, \dots , \mathbf{s}^{te}_{|\mathcal{C}_s|}, \mathbf{s}^{te}_{|\mathcal{C}_s|+1}, \dots , \mathbf{s}^{te}_{|\mathcal{C}_s|+|\mathcal{C}_u|})^\top \\ \in \mathbb{R}^{|\mathcal{C}| \times K}$.
In both cases, given a function $\hat{f}$ learned in the training phase, we want to estimate a prediction $\hat{y}$ for a visual testing sample $\mathbf{x}$ such that:
\begin{equation} \label{eq:prediction}
\hat{y} = \underset{c \in \mathcal{C}^{te}}{\text{argmax}} \hat{f}(\mathbf{x}; \mathbf{s}^{te}_c)
\end{equation}
In classical ZSL, performance is measured by the \emph{accuracy of unseen classes among unseen classes}, noted $A_{\mathcal{U} \rightarrow \mathcal{C}_u}$, while in GZSL we are interested in the \emph{accuracy of unseen classes among all classes} and the \emph{accuracy of seen classes among all classes}, noted respectively $A_{\mathcal{U} \rightarrow \mathcal{C}}$ and $A_{\mathcal{S} \rightarrow \mathcal{C}}$ as in \cite{chao2016}. $A_{\mathcal{S} \rightarrow \mathcal{C}_s}$ is similarly defined.

\subsection{Calibration and GZSL split} \label{sec:calibration}

As evidenced by~\cite{chao2016}, when a ZSL model is applied in a GZSL setting, $A_{\mathcal{S} \rightarrow \mathcal{C}}$ is usually significantly higher than $A_{\mathcal{U} \rightarrow \mathcal{C}}$.
This is because most samples from unseen classes are incorrectly classified into one of the seen classes. To address this, a \emph{calibration factor} $\gamma$ is added in~\cite{chao2016} to penalize seen classes. Eq.~(\ref{eq:prediction}) then becomes:
\begin{equation} \label{eq:calibrated_prediction}
  \hat{y} = \underset{c \in \mathcal{C}^{te}}{\text{argmax}} \left(\hat{f}(\mathbf{x}; \mathbf{s}^{te}_c) - \gamma \mathds{1}[c \in \mathcal{C}_s] \right)
\end{equation}
where $\mathds{1}[ \cdot ]$ is an indicator function.

The \emph{Accuracy Under Seen-Unseen Curve (AUSUC)} metric also proposed in~\cite{chao2016} is defined as the area under the curve representing $A_{\mathcal{S} \rightarrow \mathcal{C}}$ versus $A_{\mathcal{U} \rightarrow \mathcal{C}}$ when $\gamma$ varies from $-\infty$ to $+\infty$, which shows the trade-off between the two.

\begin{figure}
\includegraphics[width=\textwidth]{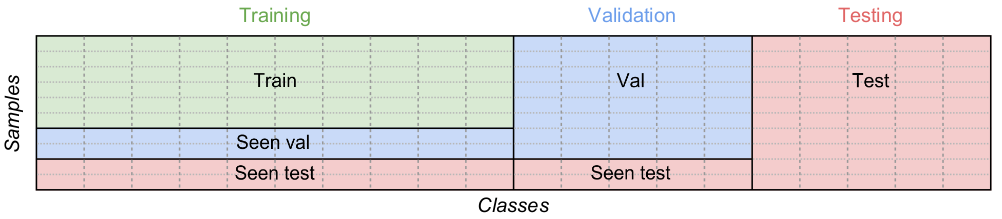}
\caption{Illustration of the different splits. Each column is a class and each cell is a sample. In this example there are 20 different classes with 10 samples per class. Five classes are used for testing, five other for validation and the remaining ten for training. Among the samples from the validation and training classes, 20\% are kept for testing  (\emph{seen test set}) and 20\% more samples from training classes are kept for validation (\emph{seen validation set}).} 
\label{fig:splits}
\end{figure}

Instead of computing a metric involving all possible trade-offs between $A_{\mathcal{U} \rightarrow \mathcal{C}}$ and $A_{\mathcal{S} \rightarrow \mathcal{C}}$, we look for a single specific value of $\gamma$, corresponding to the best compromise between the two as measured by the harmonic mean of $A_{\mathcal{U} \rightarrow \mathcal{C}}$ and $A_{\mathcal{S} \rightarrow \mathcal{C}}$~\cite{xian2017cvpr}.
We propose to determine the optimal value of $\gamma$ with a cross-validation specific to GZSL. Usually in machine learning a dataset is divided at random into three parts: a training, a validation and a testing set. In classical ZSL, this splitting process is done with respect to the classes as opposed to the samples: a set of classes is used for training, a disjoint set for validation and a final mutually disjoint set for testing. In GZSL, a fraction (usually 20\%) of the samples from the validation and training sets are kept for testing time to be used as test samples from seen classes. We refer to this set as the \emph{seen test set}. Note that here \emph{seen} only indicates that these samples belong to seen classes, not that they have been used during training. To be able to cross-validate parameters for GZSL, we further keep an additional 20\% of the remaining training set to be used as samples from seen classes when cross-validating parameters; we refer to this set as the \emph{seen validation set}. Fig.~\ref{fig:splits} illustrates this partitioning.

To determine the optimal value of $\gamma$ we first train a model on the GZSL training set. We then use the (GZSL) validation set and the seen validation set to compute the GZSL metric (the harmonic mean) and keep the value $\gamma^{*}$ that maximizes this metric. The ZSL model is subsequently re-trained on the training, validation and seen validation sets, then class similarities are computed for the test set.
The value $\gamma^{*}$ is subtracted from the similarities of seen classes and the resulting similarities are used to compute the final GZSL score.

\subsection{Regularization for GZSL} \label{sec:regularization}

\begin{figure}
\begin{tabular}{cc}
\includegraphics[width=7.2cm]{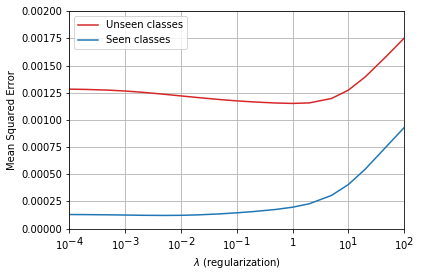}&
\includegraphics[width=4.37cm]{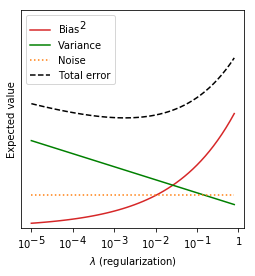}\\
(a)&(b)
\end{tabular}
\caption{(a) MSE of predicted attributes (averaged over attributes and samples) as a function of the regularization parameter $\lambda$; (b) Illustration of the bias-variance decomposition.}
\label{fig:regularization}
\end{figure}

The usual approach to optimize the regularized loss (Eq.~(\ref{eq:objective})) in GZSL consists in using the value of $\lambda$ determined on the ZSL task.
We argue here that this is unlikely to be optimal and provide some insight to justify our position. Then, we propose a simple method to determine a better value of $\lambda$ to improve performance in GZSL.

Figure \ref{fig:illustration} shows $A_{\mathcal{U} \rightarrow \mathcal{C}_u}$ and $A_{\mathcal{S} \rightarrow \mathcal{C}_s}$ as a function of $\lambda$ for a regularized linear model (ridge regression~\cite{bishop,ridge}), measured on the first validation splits of the \emph{proposed splits} of \cite{xian2017cvpr} on CUB~\cite{cub} and AwA2~\cite{xian2017arxiv}.

In each case, there is a value of $\lambda$ that maximizes the ZSL score $A_{\mathcal{U} \rightarrow \mathcal{C}_u}$, indicated by the red dotted vertical line, that we note $\lambda^{*}_{ZSL}$. 
The overall tendency for $A_{\mathcal{S} \rightarrow \mathcal{C}_s}$ is to decrease as $\lambda$ increases. This is not a concern for the ZSL task, since it only considers samples from unseen classes. However, for the GZSL task, we want the best trade-off between $A_{\mathcal{U} \rightarrow \mathcal{C}}$ and $A_{\mathcal{S} \rightarrow \mathcal{C}}$. Note that $A_{\mathcal{U} \rightarrow \mathcal{C}} \leq A_{\mathcal{U} \rightarrow \mathcal{C}_u}$ and $A_{\mathcal{S} \rightarrow \mathcal{C}} \leq A_{\mathcal{S} \rightarrow \mathcal{C}_s}$, with equality only if we are able to perfectly distinguish samples from seen and unseen classes. It follows that $\lambda^{*}_{GZSL}$, the value of $\lambda$ that maximizes the GZSL score, is not necessarily the same as $\lambda^{*}_{ZSL}$: a small decrease from $\lambda^{*}_{ZSL}$ can significantly increase $A_{\mathcal{S} \rightarrow \mathcal{C}_s}$ while only slightly penalizing $A_{\mathcal{U} \rightarrow \mathcal{C}_u}$. This has a similar impact on the maximum values obtainable by $A_{\mathcal{S} \rightarrow \mathcal{C}}$ and $A_{\mathcal{U} \rightarrow \mathcal{C}}$, and can ultimately improve the GZSL score. We quantify in Sec.~\ref{sec:results} the gains attributed to the use of $\lambda^{*}_{GZSL}$.

The reason why $\lambda$ affects $A_{\mathcal{U} \rightarrow \mathcal{C}_u}$ and $A_{\mathcal{S} \rightarrow \mathcal{C}_s}$ in this way can be explained with the bias-variance decomposition. For regression, we generally assume that we are given a dataset $\mathcal{D} = (\mathbf{X}, \mathbf{t})$, with samples $(\mathbf{x}_n, t_n)$ independently drawn from a joint distribution $p(\mathbf{x}, t)$, such that $p(t|\mathbf{x}) = \mathcal{N}(t|h(\mathbf{x}), \sigma^2)$, where $h$ is the true dependence.
For a prediction function $\hat{h}$ estimated from $\mathcal{D}$ we can then write the expected loss on a new pair $(\mathbf{x}, t)$ as:
\begin{equation} \label{eq:bias_variance}
\mathbb{E}_{\mathcal{D}, \mathbf{x}, t}[(t - \hat{h}(\mathbf{x}))^2] = \sigma^2 + \mathbb{E}_{\mathcal{D}, \mathbf{x}}[(h(\mathbf{x}) - \hat{h}(\mathbf{x}))^2] + \text{var}_{\mathcal{D}, \mathbf{x}}[\hat{h}(\mathbf{x})]
\end{equation}
where the first term is the intrinsic noise of the dataset, the second is the (squared) bias of the predictor and the third is the variance in the estimation of the predictor. It can be shown~\cite{bishop,ridge} that for ridge regression the bias increases and the variance decreases with the regularization parameter $\lambda$, as illustrated in Fig.~\ref{fig:regularization}(b).

In the case of ZSL, $\mathbf{x}$ corresponds to visual samples and $t$ to attribute(s) to be estimated from $\mathbf{x}$. The variance comes from both the differences between samples from the same class (intra-class variance) and from the differences between classes (inter-class variance). Intra-class variance is usually significantly smaller than inter-class variance in ZSL.
Therefore, most of the variance in Eq.~(\ref{eq:bias_variance}) can be attributed to the choice of training classes $\mathcal{C}_s$. For samples from unseen classes, the bias-variance decomposition applies and there exists a $\lambda$ corresponding to the best trade-off between the two. This is evidenced in Fig.~\ref{fig:regularization}(a), where the red curve shows the Mean Squared Error (MSE) in the predictions of attributes from unseen classes as a function of $\lambda$, for a regularized linear model on the first validation split of AwA2~\cite{xian2017arxiv}.

For a sample from a seen class, the variance attributable to the choice of the training classes is much smaller since, by definition, the seen class is present in the training dataset. This allows to better estimate attributes from seen classes and most of the expected error therefore comes from the intrinsic noise and the bias. Thus, the expected error mostly increases with $\lambda$, as evidenced by the blue curve in Fig.~\ref{fig:regularization}(a). 
If we plausibly assume that the accuracy of predictions for samples from a given class depends on how well we estimate their attributes, this explains both why predictions are better for samples from seen classes than from unseen classes and why their behavior with respect to $\lambda$ is different.

We then suggest the following procedure to select the optimal value of $\lambda$: we repeat the protocol described in Sec.~\ref{sec:calibration} for selecting $\gamma^{*}$ and we take the value of $\lambda$ which gives the best result for the harmonic mean between $A_{\mathcal{U} \rightarrow \mathcal{C}}$ and $A_{\mathcal{S} \rightarrow \mathcal{C}}$ on the validation set \emph{after} having subtracted $\gamma^{*}$ from the similarities of seen classes. The rest of the process is identical: we retrain the ZSL model on the training, validation and seen validation sets with the hyperparameter $\lambda^{*}_{GZSL}$ that we just determined, we compute similarities for the test set, subtract $\gamma^{*}$ from the similarities of seen classes and compute the resulting GZSL score.

\section{Experimental evaluation}

\subsection{Methods}

We independently reimplemented six methods frequently cited in the literature to evaluate them with our protocol: ALE~\cite{ale}, DeViSE~\cite{devise}, SJE~\cite{sje}, Sync~\cite{sync}, ESZSL~\cite{eszsl} and SAE~\cite{sae}.

In addition, we also evaluate two simple linear models. 
Linear\textsubscript{$\mathcal{V} \rightarrow \mathcal{S}$} applies a linear mapping $\mathbf{W} \in \mathbb{R}^{K \times D}$ from the visual space $\mathcal{V}$ to the semantic space $\mathcal{S}$ to minimize standard MSE. With $\mathbf{T}^{tr} = (\mathbf{s}^{tr}_{y^{tr}_1}, ... , \mathbf{s}^{tr}_{y^{tr}_{N}})^\top \in \mathbb{R}^{N^{tr} \times K}$ the matrix whose rows correspond to the class prototypes associated to each training sample based on its label, the loss function can be formulated as:
\begin{equation} \label{eq:loss_lvs}
\frac{1}{N^{tr}} \lVert \mathbf{X}^{tr} \mathbf{W}^\top - \mathbf{T}^{tr} \rVert^2_{F} + \lambda \lVert \mathbf{W} \rVert^2_{F}
\end{equation}
%
%
Linear\textsubscript{$\mathcal{S} \rightarrow \mathcal{V}$} is based on \cite{lsv} where the authors argue that using the semantic space as the embedding space reduces the variance of the projected points and thus aggravates the hubness problem~\cite{radovanovic}. They suggest instead to project semantic class prototypes onto the visual space and to compute similarities in this space.
Keeping $\mathbf{W} \in \mathbb{R}^{K \times D}$ as our linear mapping, we formulate the loss function as:
\begin{equation} \label{eq:loss_lsv}
\frac{1}{N^{tr}} \lVert \mathbf{X}^{tr} - \mathbf{T}^{tr} \mathbf{W} \rVert^2_{F} + \lambda \lVert \mathbf{W} \rVert^2_{F}
\end{equation}

We can easily obtain closed-form solutions for the two models from the objective functions~(\ref{eq:loss_lvs}) and (\ref{eq:loss_lsv}). For the Linear\textsubscript{$\mathcal{V} \rightarrow \mathcal{S}$} model we have $\mathbf{W} = {\mathbf{T}^{tr}}^\top \mathbf{X}^{tr} ({\mathbf{X}^{tr}}^\top \mathbf{X}^{tr} + \lambda N^{tr} \mathbf{I}_D)^{-1} \inlineeqnum$
and for the Linear\textsubscript{$\mathcal{S} \rightarrow \mathcal{V}$} model $\mathbf{W} = ({\mathbf{T}^{tr}}^\top \mathbf{T}^{tr} + \lambda N^{tr} \mathbf{I}_K)^{-1} {\mathbf{T}^{tr}}^\top {\mathbf{X}^{tr}} \inlineeqnum$. 

\subsection{Experimental setting} \label{sec:splits}

\subsubsection{Datasets} We perform our experiments on two standard datasets for ZSL: Caltech-UCSD-Birds 200-2011 (CUB)~\cite{cub} and Animals with Attributes2\footnote{AwA2 was recently proposed in~\cite{xian2017arxiv} as a replacement for the Animals with Attributes (AwA) dataset~\cite{awa} whose images are not publicly available.}
(AwA2) \cite{xian2017arxiv}. 
Results for two additional datasets (SUN Attribute Database~\cite{sun} and Attributes Pascal and Yahoo~\cite{apy}) are available in the supplementary material. 
CUB is a fine-grained dataset composed of 11788 pictures of birds from 200 species (\emph{black footed albatross}, \dots, \emph{common yellowthroat}). 
It comes with 312-dimensional binary attributes for each picture, that are averaged by class to obtain semantic class prototypes.
AwA2 is a coarse-grained dataset comprising 37322 pictures of 50 animal species (\emph{antelope}, \dots, \emph{zebra}). For each class, 85-dimensional attributes  are provided.

\subsubsection{Splits} The best performing ZSL methods usually rely on visual features obtained with deep neural networks pre-trained on ImageNet~\cite{imagenet}, such as GoogLeNet \cite{googlenet} or ResNet~\cite{resnet}. 
As evidenced by~\cite{xian2017cvpr}, this induces a huge bias for ZSL datasets whose classes are not disjoint from categories of ImageNet, as is the case with AwA2, since test classes cannot be considered truly unseen. We therefore adopt the approach of \cite{xian2017cvpr} 
and use their terms \emph{Standard Split} (S.S.) for the split widely used in the literature and \emph{Proposed Split} (P.S.) for the split they introduce. 
The training and validation splits are further divided for GZSL as described in Sec.~\ref{sec:calibration}.
Statistics regarding the (GZSL) splits are given in the supplementary material.

\subsubsection{Settings} Attributes are normalized such that each class prototype has unit $\ell 2$ norm. We use the 101-layered ResNet~\cite{resnet} pre-trained on ImageNet~\cite{imagenet} as visual features extractor, keeping the $D = 2048$ activations of the last pooling units.

\subsubsection{Metrics} For ZSL, we evaluate the accuracy of samples from unseen classes among unseen classes $A_{\mathcal{U} \rightarrow \mathcal{C}_u}$. There are two possible ways to define accuracy: most of the literature uses \emph{per sample} accuracy, defined as $100 \cdot \frac{1}{N^{te}} \sum_{n=1}^{N^{te}} \mathds{1}[\hat{y}(\mathbf{x}_n^{te}) = y_n^{te}]$, while in~\cite{xian2017cvpr} it is argued that \emph{per class} accuracy, defined as \\ $100 \cdot \frac{1}{|\mathcal{C}^{te}|}  \sum_{c \in \mathcal{C}^{te}} \frac{1}{|\{n | y_n = c\}|} \sum_{\substack{n \\ y_n = c}} \mathds{1}[\hat{y}(\mathbf{x}_n^{te}) = y_n^{te}]$, better takes class imbalance into account.

We report per class accuracy for fair comparison with the extensive results of~\cite{xian2017cvpr}. Nonetheless, to enable comparison with the rest of the literature, we also provide per sample accuracy results in Table~\ref{table:10crop}.
For GZSL we compute the harmonic mean between $A_{\mathcal{U} \rightarrow \mathcal{C}}$ and $A_{\mathcal{S} \rightarrow \mathcal{C}}$, defined as $\frac{2 \cdot A_{\mathcal{U} \rightarrow \mathcal{C}} \cdot A_{\mathcal{S} \rightarrow \mathcal{C}}}{A_{\mathcal{U} \rightarrow \mathcal{C}} + A_{\mathcal{S} \rightarrow \mathcal{C}}}$.

Accuracy is again assumed to be per class unless otherwise stated.

\subsection{Results} \label{sec:results}

\begin{table}[t]
\caption{ZSL score: per-class accuracy $A_{\mathcal{U} \rightarrow \mathcal{C}_u}$, as reported in \cite{xian2017cvpr} and independently reproduced. S.S.: Standard Split, P.S.: Proposed Split~\cite{xian2017cvpr}. Averaged over 5 runs.}
\label{table:zsl}
\begin{center}
\resizebox{\textwidth}{!}{%
\begin{tabular}{|l||c c|c c||c c|c c|}
\hline
\multirow{3}{*}{\textbf{Method}} & \multicolumn{4}{|c||}{\textbf{CUB}~\cite{cub}} & \multicolumn{4}{|c|}{\textbf{AwA2}~\cite{xian2017arxiv}} \\
\cline{2-9}
& \multicolumn{2}{|c|}{\textbf{Reported in}~\cite{xian2017cvpr}} & \multicolumn{2}{|c||}{\textbf{Reproduced}} & \multicolumn{2}{|c|}{\textbf{Reported in}~\cite{xian2017cvpr}} & \multicolumn{2}{|c|}{\textbf{Reproduced}} \\
& \textbf{S.S.} & \textbf{P.S.} & \textbf{S.S.} & \textbf{P.S.} & \textbf{S.S.} & \textbf{P.S.} & \textbf{S.S.} & \textbf{P.S.} \\
\hline
Linear\textsubscript{$\mathcal{V} \rightarrow \mathcal{S}$} & \emph{n/a} & \emph{n/a} & 41.0 $\pm$ 0.0 & 41.8 $\pm$ 0.0 &
\emph{n/a} & \emph{n/a} & 68.2 $\pm$ 0.0 & 49.7 $\pm$ 0.0 \\
Linear\textsubscript{$\mathcal{S} \rightarrow \mathcal{V}$} & \emph{n/a} & \emph{n/a} & 56.0 $\pm$ 0.0 & 53.5 $\pm$ 0.0 &
\emph{n/a} & \emph{n/a} & \textbf{85.5} $\pm$ 0.0 & \textbf{68.9} $\pm$ 0.0 \\
ALE~\cite{ale} & 53.2 & 54.9 & 54.8 $\pm$ 0.8 & 54.0 $\pm$ 1.2 &
80.3 & \textbf{62.5} & 80.3 $\pm$ 2.2 & 62.9 $\pm$ 2.3 \\
DeViSE~\cite{devise} & 53.2 & 52.0 & 52.5 $\pm$ 0.9 & 52.6 $\pm$ 1.3 &
68.6 & 59.7 & 76.6 $\pm$ 1.6 & 62.1 $\pm$ 1.6 \\
SJE~\cite{sje} & \textbf{55.3} & 53.9 & 53.8 $\pm$ 2.3 & 49.2 $\pm$ 1.4 &
69.5 & 61.9 & 80.4 $\pm$ 2.9 & 62.2 $\pm$ 1.2 \\
ESZSL~\cite{eszsl} & 55.1 & 53.9 & 34.9 $\pm$ 0.0 & 34.9 $\pm$ 0.0 &
75.6 & 58.6 & 70.5 $\pm$ 0.0 & 50.8 $\pm$ 0.0 \\
Sync~\cite{sync} & 54.1 & \textbf{55.6} & \textbf{56.4} $\pm$ 0.9 & \textbf{54.8} $\pm$ 0.6 &
71.2 & 46.6 & 65.6 $\pm$ 0.8 & 58.1 $\pm$ 0.8 \\
SAE~\cite{sae} & 33.4 & 33.3 & 56.2 $\pm$ 0.0 & 53.3 $\pm$ 0.0 &
\textbf{80.7} & 54.1 & 81.1 $\pm$ 0.0 & 62.8 $\pm$ 0.0 \\
\hline
\end{tabular}
}
\end{center}
\end{table}

\begin{table}
\caption{GZSL score (harmonic mean of $A_{\mathcal{U} \rightarrow \mathcal{C}}$ and $A_{\mathcal{S} \rightarrow \mathcal{C}}$, per class accuracy) with and without calibration and GZSL regularization. On Proposed Split~\cite{xian2017cvpr}, averaged over 5 runs.}
\label{table:gzsl}
\begin{center}
\resizebox{\textwidth}{!}{%
\begin{tabular}{|l||c|c c c||c|c c c|}
\hline
\multirow{2}{*}{\textbf{Method}} & \multicolumn{4}{|c||}{\textbf{CUB}~\cite{cub}} & \multicolumn{4}{|c|}{\textbf{AwA2}~\cite{xian2017arxiv}} \\
\cline{2-9}
& \textbf{Reported in}~\cite{xian2017cvpr} & \multicolumn{3}{|c||}{\textbf{Ours}} &
\textbf{Reported in}~\cite{xian2017cvpr} & \multicolumn{3}{|c|}{\textbf{Ours}} \\
\hline
with \textbf{calibration} & - & - & \cmark & \cmark &
- & - & \cmark & \cmark \\
with $\boldsymbol{\lambda}^{*}_{\mathbf{GZSL}}$ & - & - & - & \cmark &
- & - & - & \cmark \\
\hline
Linear\textsubscript{$\mathcal{V} \rightarrow \mathcal{S}$} & \emph{n/a} & 18.2 & 34.3 & 35.5 &
\emph{n/a} & 8.3 & 47.3 & 48.1 \\
Linear\textsubscript{$\mathcal{S} \rightarrow \mathcal{V}$} & \emph{n/a} & 32.5 & 41.9 & 43.5 &
\emph{n/a} & 44.3 & 62.7 & 64.0 \\
ALE~\cite{ale} & 34.4 & 35.6 & 45.1 & 46.2 &
23.9 & 26.9 & 55.8 & 55.8 \\
DeViSE~\cite{devise} & 32.8 & 35.1 & 43.6 & 43.4 &
27.8 & 17.4 & 54.6 & 54.6 \\
SJE~\cite{sje} & 33.6 & 29.7 & 41.2 & 44.2 &
14.4 & 28.9 & 58.2 & 59.0 \\
ESZSL~\cite{eszsl} & 21.0 & 17.9 & 33.7 & 33.9 &
11.0 & 39.9 & 53.6 & 53.7 \\
Sync~\cite{sync} & 19.8 & 33.2 & 46.2 & 47.6 &
18.0 & 30.6 & 61.0 & 61.0 \\
SAE~\cite{sae} & 13.6 & 25.7 & 43.1 & 43.1 &
2.2 & 29.5 & 60.2 & 60.2 \\
\hline
Average & 25.9 & 28.5 & 41.1  & \textbf{42.2} & 
16.2 & 28.2 & 56.7 & \textbf{57.1} \\
\hline
\end{tabular}
}
\end{center}
\end{table}

\begin{table}[t]
\caption{ZSL and GZSL scores with 10-crop features, evaluated with per class (p.c.) and per sample (p.s.) accuracies. With calibration and $\lambda^{*}_{GZSL}$. On P.S.~\cite{xian2017cvpr}, averaged over 5 runs.}
\label{table:10crop}
\begin{center}
\resizebox{\textwidth}{!}{%
\begin{tabular}{|l||c c|c c||c c|c c|}
\hline
\multirow{3}{*}{\textbf{Method}} & \multicolumn{4}{|c||}{\textbf{CUB}~\cite{cub}} & \multicolumn{4}{|c|}{\textbf{AwA2}~\cite{xian2017arxiv}} \\
\cline{2-9}
& \multicolumn{2}{|c|}{\textbf{ZSL}} & \multicolumn{2}{|c||}{\textbf{GZSL}} &
\multicolumn{2}{|c|}{\textbf{ZSL}} & \multicolumn{2}{|c|}{\textbf{GZSL}} \\
& \textbf{Acc. p.c.} & \textbf{Acc. p.s.} & \textbf{H. p.c.} & \textbf{H. p.s.} &
\textbf{Acc. p.c.} & \textbf{Acc. p.s.} & \textbf{H. p.c.} & \textbf{H. p.s.} \\
\hline
Linear\textsubscript{$\mathcal{V} \rightarrow \mathcal{S}$} & 45.6 & 45.6 & 39.8 & 39.8 &
51.0 & 43.6 & 49.0 & 45.6 \\
Linear\textsubscript{$\mathcal{S} \rightarrow \mathcal{V}$} & 57.1 & 57.2 & 47.7 & 48.0 &
\textbf{70.4} & \textbf{69.3} & \textbf{65.1} & \textbf{68.7} \\
ALE~\cite{ale} & 57.4 & 57.5 & \textbf{49.2} & \textbf{49.3} &
63.0 & 61.1 & 56.9 & 55.5 \\
DeViSE~\cite{devise} & 52.9 & 52.9 & 42.4 & 42.5 &
63.1 & 62.2 & 55.0 & 50.6 \\
SJE~\cite{sje} & 51.9 & 52.1 & 46.7 & 46.9 &
63.8 & 61.6 & 59.4 & 57.6 \\
ESZSL~\cite{eszsl} & 39.0 & 38.8 & 38.7 & 38.6 &
52.6 & 51.9 & 54.4 & 57.9 \\
Sync~\cite{sync} & \textbf{57.5} & \textbf{57.6} & 48.9 & 49.1 &
59.3 & 56.1 & 62.6 & 63.2 \\
SAE~\cite{sae} & 56.1 & 56.2 & 46.3 & 46.6 &
63.5 & 65.4 & 62.3 & 63.6 \\
\hline
\end{tabular}
}
\end{center}
\end{table}

We first evaluate the performances of the different methods in a classical ZSL setting. Table~\ref{table:zsl} shows the average per class accuracy measured on testing sets of the Standard Splits (S.S.) and the Proposed Splits (P.S.)~\cite{xian2017cvpr} of CUB~\cite{cub} and AwA2~\cite{xian2017arxiv}. We report the average score and the standard deviation over 5 runs with different random initializations.
We also report the results from \cite{xian2017cvpr}.
We see that some methods such as SJE~\cite{sje} have high variability with respect to the initialization; for such methods, it is good practice to report average results since a single test run may not be representative of the true performance of the model. On the other hand, methods with closed-form or deterministic solutions such as the Linear\textsubscript{$\mathcal{V} \rightarrow \mathcal{S}$}, Linear\textsubscript{$\mathcal{S} \rightarrow \mathcal{V}$}, ESZSL~\cite{eszsl} or SAE~\cite{sae} are not dependent on the initialization and thus have a standard deviation of 0.

Most of the reproduced scores are consistent with \cite{xian2017cvpr}, with two notable exceptions: first, a significant increase in performance is observed with SAE~\cite{sae} and can be explained by the fact that similarities are computed in the visual space, with results close to those of the Linear\textsubscript{$\mathcal{S} \rightarrow \mathcal{V}$} model (results are close to those of Linear\textsubscript{$\mathcal{V} \rightarrow \mathcal{S}$} when similarities are computed in the semantic space).
Second, the score for ESZSL~\cite{eszsl} is significantly lower than reported in \cite{xian2017cvpr}. We found that the use of non-normalized attributes enables to reach performances comparable with \cite{xian2017cvpr}, but we could not reproduce the reported results for ESZSL~\cite{eszsl} with normalized attributes. For the sake of consistency, we chose to report results obtained with normalized attributes.

Table~\ref{table:gzsl} shows results for GZSL. We measure the harmonic mean between per class accuracies $A_{\mathcal{U} \rightarrow \mathcal{C}}$ and $A_{\mathcal{S} \rightarrow \mathcal{C}}$ on the testing set of the Proposed Split~\cite{xian2017cvpr}.
We evaluate three settings: a ZSL model applied directly in a GZSL setting, i.e. with no calibration and a regularization specific to ZSL ($\lambda^{*}_{ZSL}$) as opposed to GZSL ($\lambda^{*}_{GZSL}$); a ZSL model with calibration and ZSL regularization $\lambda^{*}_{ZSL}$; and a ZSL model with calibration and regularization $\lambda^{*}_{GZSL}$ specific to the GZSL problem. We report the average score over 5 runs; standard deviations are available in the supplementary material.
We also report the results from \cite{xian2017cvpr}, which correspond to the setting with no calibration and no $\lambda^{*}_{GZSL}$.
We can see that the calibration process significantly improves GZSL performance: in our experiments, the average score for all models improves from 28.5 with no calibration to 41.1 with calibration on CUB, and from 28.2 to 56.7 on AwA2. It is worth noting that the lowest score with calibration is close to or higher than the highest score without.
The use of a regularization parameter specific to the GZSL task can lead to an additional improvement in performance. In some cases, the optimal $\lambda$ is the same for the ZSL task and the GZSL task on the validation set, leading to no additional improvement over the score with calibration. However, every time they are different, $\lambda^{*}_{GZSL}$ is smaller than $\lambda^{*}_{ZSL}$, as expected from the results in Sec.~\ref{sec:regularization}. The only exception is with DeViSE~\cite{devise} on CUB: a $\lambda^{*}_{GZSL}$ higher than $\lambda^{*}_{ZSL}$ was selected during cross-validation, probably due to random noise, resulting in a slightly lower final GZSL score.

Table~\ref{table:10crop} shows results with improved visual features; each original $256 \times 256$ image is cropped into ten $224 \times 224$ images: one in each corner and one in the center for both the original image and its horizontal symmetry. The ResNet features of the resulting images are averaged to obtain a 2048-dimensional vector. We report results for ZSL ($A_{\mathcal{C} \rightarrow \mathcal{C}_u}$, abbreviated \emph{Acc.}) and GZSL (using the harmonic mean metric, abbreviated \emph{H.})
on the testing set of the Proposed Split~\cite{xian2017cvpr}. In order to facilitate fair comparison with the rest of the literature, both per class \emph{(p.c.)} and per sample \emph{(p.s.)} metrics are reported. Additional metrics like AUSUC\cite{chao2016} are available in the supplementary material.
Results with 10-cropped visual features are almost always better than the results with standard visual features in Table~\ref{table:gzsl}. The per sample metrics are on average not very different from the per class metrics. This is not surprising since classes in both CUB and AwA2 are fairly balanced.

\section{Conclusion}

We proposed a simple process for applying ZSL methods in a GZSL setting. This process is based on the empirical observation that ZSL models perform differently on samples from seen and unseen classes. We provided insights about why this should be expected and suggested steps to overcome these problems.
Through extensive experiments, we showed that this process enables significant improvements in performance for many existing ZSL methods. Finally, we provided results under optimal conditions for these methods with different metrics to support fair comparison with the rest of the state-of-the-art.

\bibliographystyle{ieeetr}
\bibliography{bibliography_full_abbr.bib}

\clearpage

\section{Supplementary Material -- From Classical to Generalized Zero-Shot Learning: a Simple Adaptation Process}

\subsection{Details regarding the compared methods} \label{sec:sup_methods}

ALE~\cite{ale}, DeViSE~\cite{devise} and SJE~\cite{sje} employ bilinear similarity functions of the form $f(\mathbf{x}; \mathbf{s}) = \mathbf{s^\top W x}$ with $\mathbf{W} \in \mathbb{R}^{K \times D}$ combined with a hinge rank loss of the form $\text{max}(0, M + f(\mathbf{x}_n; \mathbf{s}_c) - f(\mathbf{x}_n; \mathbf{s}_{y_n}))$ for $c \neq y_n$, where $M$ is a margin usually set to a fraction of the expected value of $f(\mathbf{x}; \mathbf{s})$. 
For a given visual sample $\mathbf{x}_n$, DeViSE simply sums the hinge range losses for all values of $c \in \mathcal{C}_s$ such that $c\neq y_n$. ALE sums these terms and adds a multiplicative factor to lessen the weight of samples for which many classes contribute to the hinge rank loss. SJE only keeps the loss of the class $c$ corresponding to the maximum of the hinge loss.
We add a regularization term of the form $\lambda \lVert \mathbf{W} \rVert^2_{F}$ to all these models, $\lVert \cdot \rVert_{F}$ being the Frobenius norm. 

Sync~\cite{sync} uses \emph{"phantom" objects} located in both the visual and semantic spaces to generate classifiers for unseen classes. We employ the \emph{structured loss} described in Sec.~3.2 of \cite{sync} as it yields better results. 
ESZSL~\cite{eszsl} is a simple linear model regularized with respect to the projected visual features of class prototypes, the projected semantic features of the visual samples and the projection matrix. One of its advantages is the existence of a closed form solution. 
SAE~\cite{sae} learns an autoencoder on visual features aiming to balance a reconstruction loss and a representation loss so that its internal representation corresponds to the semantic attributes of visual features.

\subsection{Details regarding datasets and splits}

In addition to CUB~\cite{cub} and AwA2~\cite{xian2017arxiv}, we provide results for two other datasets, the SUN Attribute Database (SUN)~\cite{sun} and Attributes Pascal and Yahoo (aPY)~\cite{apy}. SUN is a fine-grained dataset comprising 717 categories of scenes (\emph{abbey}, \dots, \emph{zoo}) with 20 images per category, for a total of 14340 images. It comes with 102-dimensional vectors of attributes for each picture; they are averaged by class to obtain semantic class prototypes. aPY is a coarse-grained dataset composed of 15339 images from 3 broad categories (animals, objects and vehicles), further divided into a total of 32 subcategories (\emph{aeroplane}, \dots, \emph{zebra}). It is rather imbalanced, with the category \emph{person} representing a third of all samples. Moreover, with 21 out of 32 classes present in the ImageNet~\cite{imagenet} dataset, it is also highly biased when used in a ZSL context with deep networks pre-trained on ImageNet~\cite{imagenet} as visual features extractors. 
\cite{xian2017cvpr} also provides Proposed Splits in addition to the Standard Splits for these two other datasets; we use the same protocol as with CUB~\cite{cub} and AwA2~\cite{xian2017arxiv}, described in Sec.~3.2 of the main text, to further divide these splits for GZSL.

We thus have a total of four datasets, each with two different testing sets, with three different validation sets for each testing set. Statistics regarding these splits are available in Table~\ref{table:datasets}. For a given dataset and testing set, the average score over the three different validation splits is employed for selecting hyperparameters by cross-validation.

Intra-class and inter-class variances for each dataset are reported in Table~\ref{tab:variance}.

\begin{table}
\centering
\caption{Statistics of the dataset splits for each of the four datasets, with two testing splits (Standard Splits or S.S. and Proposed Splits or P.S. as defined in \cite{xian2017cvpr}) per dataset and three validation splits per testing split. Each cell contains a number of samples on the left and the number of corresponding classes on the right.}
\label{table:datasets}
\resizebox{\textwidth}{!}{%
\begin{tabular}{|r|c|c c c|c c|c|}
\hline
\multirow{2}{*}{\textbf{Split}} & \multirow{2}{*}{\textbf{Total}} & \multicolumn{3}{|c|}{\textbf{Training}} & \multicolumn{2}{|c|}{\textbf{Validation}} & \textbf{Testing} \\
 &  & \textbf{Train} & \textbf{Seen val} & \textbf{Seen test} & \textbf{Val} & \textbf{Seen test} & \textbf{Test}\\
\hline\hline
\multicolumn{8}{|c|}{\textbf{CUB}~\cite{cub}} \\
\hline
\textbf{S.S. val1} & \multirow{6}{*}{11788 / \emph{200}} & 3773 / \emph{100} & 943 / \emph{100} & 1178 / \emph{100} & 2369 / \emph{50} & 592 / \emph{50} & \multirow{3}{*}{2933 / \emph{50}} \\
\textbf{S.S. val2} &  & 3798 / \emph{100} & 949 / \emph{100} & 1186 / \emph{100} & 2338 / \emph{50} & 584 / \emph{50} &  \\
\textbf{S.S. val3} &  & 3784 / \emph{100} & 946 / \emph{100} & 1182 / \emph{100} & 2355 / \emph{50} & 588 / \emph{50} & \\
\cline{1-1}\cline{3-8}
\textbf{P.S. val1} &  & 3760 / \emph{100} & 940 / \emph{100} & 1175 / \emph{100} & 2357 / \emph{50} & 589 / \emph{50} & \multirow{3}{*}{2967 / \emph{50}} \\
\textbf{P.S. val2} &  & 3760 / \emph{100} & 940 / \emph{100} & 1175 / \emph{100} & 2357 / \emph{50} & 589 / \emph{50} & \\
\textbf{P.S. val3} &  & 3806 / \emph{100} & 951 / \emph{100} & 1189 / \emph{100} & 2300 / \emph{50} & 575 / \emph{50} &  \\
\hline\hline
\multicolumn{8}{|c|}{\textbf{AwA2}~\cite{xian2017arxiv}} \\
\hline
\textbf{S.S. val1} & \multirow{6}{*}{37322 / \emph{50}} & 13284 / \emph{27} & 3321 / \emph{27} & 4151 / \emph{27} & 7665 / \emph{13} & 1916 / \emph{13} & \multirow{3}{*}{6985 / \emph{10}}\\
\textbf{S.S. val2} &  & 13041 / \emph{27} & 3260 / \emph{27} & 4075 / \emph{27} & 7969 / \emph{13} & 1992 / \emph{13} & \\
\textbf{S.S. val3} & & 13908 / \emph{27} & 3476 / \emph{27} & 4345 / \emph{27} & 6887 / \emph{13} & 1721 / \emph{13} & \\
\cline{1-1}\cline{3-8}
\textbf{P.S. val1} &  & 12940 / \emph{27} &  3235 / \emph{27} & 4043 / \emph{27} & 7353 / \emph{13} & 1838 / \emph{13} & \multirow{3}{*}{7913 / \emph{10}} \\ 
\textbf{P.S. val2} &  & 13404 / \emph{27} & 3351 / \emph{27} & 4188 / \emph{27} & 6773 / \emph{13} & 1693 / \emph{13} & \\ 
\textbf{P.S. val3} &  & 12965 / \emph{27} & 3241 / \emph{27} & 4051 / \emph{27} & 7322 / \emph{13} & 1830 / \emph{13} & \\ 
\hline\hline
\multicolumn{8}{|c|}{\textbf{SUN}~\cite{sun}} \\
\hline
\textbf{S.S. val1} & \multirow{6}{*}{14340 / \emph{717}} & 7437 / \emph{581} & 1859 / \emph{581} & 2324 / \emph{581} & 1024 / \emph{64} & 256 / \emph{64} & \multirow{3}{*}{1440 / \emph{72}}\\
\textbf{S.S. val2} &  & 7437 / \emph{581} & 1859 / \emph{581} & 2324 / \emph{581} & 1024 / \emph{64} & 256 / \emph{64} & \\
\textbf{S.S. val3} & & 7437 / \emph{581} & 1859 / \emph{581} & 2324 / \emph{581} & 1024 / \emph{64} & 256 / \emph{64} & \\
\cline{1-1}\cline{3-8}
\textbf{P.S. val1} &  & 7424 / \emph{580} & 1856 / \emph{580} & 2320 / \emph{580} & 1040 / \emph{65} & 260 / \emph{65} & \multirow{3}{*}{1440 / \emph{72}} \\ 
\textbf{P.S. val2} &  & 7424 / \emph{580} & 1856 / \emph{580} & 2320 / \emph{580} & 1040 / \emph{65} & 260 / \emph{65} & \\ 
\textbf{P.S. val3} &  & 7424 / \emph{580} & 1856 / \emph{580} & 2320 / \emph{580} & 1040 / \emph{65} & 260 / \emph{65} & \\ 
\hline\hline
\multicolumn{8}{|c|}{\textbf{aPY}~\cite{apy}} \\
\hline
\textbf{S.S. val1} & \multirow{6}{*}{15339 / \emph{32}} & 6424 / \emph{13} & 1606 / \emph{13} & 2007 / \emph{13} & 2127 / \emph{7} & 531 / \emph{7} & \multirow{3}{*}{2644 / \emph{12}}\\
\textbf{S.S. val2} &  & 6608 / \emph{13} & 1652 / \emph{13} & 2064 / \emph{13} & 1897 / \emph{7} & 474 / \emph{7} & \\
\textbf{S.S. val3} & & 6644 / \emph{13} & 1660 / \emph{13} & 2076 / \emph{13} & 1852 / \emph{7} & 463 / \emph{7} & \\
\cline{1-1}\cline{3-8}
\textbf{P.S. val1} &  & 3896 / \emph{15} & 973 / \emph{15} & 1217 / \emph{15} & 1064 / \emph{5} & 265 / \emph{5} & \multirow{3}{*}{7924 / \emph{12}} \\ 
\textbf{P.S. val2} &  & 3716 / \emph{15} & 928 / \emph{15} & 1161 / \emph{15} & 1288 / \emph{5} & 322 / \emph{5} & \\ 
\textbf{P.S. val3} &  & 3271 / \emph{15} & 817 / \emph{15} & 1021 / \emph{15} & 1845 / \emph{5} & 461 / \emph{5} & \\ 
\hline
\end{tabular}
}
\end{table}

\begin{table}
\centering
\caption{Intra-class and inter-class variance for each dataset.}
\label{tab:variance}
\begin{tabular}{|l|c|c|}
\hline
\textbf{Dataset} & \textbf{Intra-class variance} & \textbf{Inter-class variance} \\
\hline
\textbf{CUB~\cite{cub}} & 138.0 & 231.9 \\ \hline
\textbf{AwA2~\cite{xian2017arxiv}} & 226.4 & 379.0 \\ \hline
\textbf{SUN~\cite{sun}} & 239.9 & 397.3 \\ \hline
\textbf{aPY~\cite{apy}} & 262.4 & 370.5 \\ \hline
\end{tabular}
\end{table}

\subsection{Additional results}

Table~\ref{table:sup_zsl} is an extended version of Table~1 in the main text. It reports the per class accuracy $A_{\mathcal{U} \rightarrow \mathcal{C}_u}$ measured on the Standard Splits (S.S.) and the Proposed Splits (P.S.)~\cite{xian2017cvpr} of CUB~\cite{cub} and AwA2~\cite{xian2017arxiv}, with the addition of the two datasets SUN~\cite{sun} and aPY~\cite{apy}. Results are averaged over five runs with different random initializations.

Results on SUN~\cite{sun} are mostly consistent with what has been observed on CUB~\cite{cub} and AwA2~\cite{xian2017arxiv}. The addition of a regularization parameter to ALE~\cite{ale}, DeViSE~\cite{devise} and SJE~\cite{sje} enables a slight increase in the ZSL score. 
The Linear\textsubscript{$\mathcal{S} \rightarrow \mathcal{V}$} model reaches performances close to or even better than the state-of-the-art. 
A significant increase in performance is observed with SAE~\cite{sae} and can be explained by the fact that similarities are computed in the visual space, with results close to those of the Linear\textsubscript{$\mathcal{S} \rightarrow \mathcal{V}$} model (results are close to those of Linear\textsubscript{$\mathcal{V} \rightarrow \mathcal{S}$} when similarities are computed in the semantic space).
The score for ESZSL~\cite{eszsl} is significantly lower than reported in \cite{xian2017cvpr}. We found that the use of non-normalized attributes enables to reach performances comparable with \cite{xian2017cvpr}, but we could not reproduce the reported results for ESZSL~\cite{eszsl} with normalized attributes. 
For the sake of consistency, we chose to report results obtained with normalized attributes.

Results on aPY~\cite{apy} are more mixed: important differences between reported and reproduced scores exist on at least one split for most methods. This can be explained by the nature of this dataset: the low number of classes, the class imbalance and the high bias due to the presence of most of its classes in ImageNet~\cite{imagenet} imply that the validation sets can be significantly different from the test sets. For example, the average validation score on the three validation splits is 55.2 for DeViSE~\cite{devise}, but this score drops to an average of 29.0 on the testing set. It is therefore difficult to find relevant hyperparameters.

Table~\ref{table:sup_gzsl} is an extended version of Table~2 in the main text. It reports the average GZSL score, as measured by the harmonic mean between $A_{\mathcal{U} \rightarrow \mathcal{C}}$ and $A_{\mathcal{S} \rightarrow \mathcal{C}}$, for the evaluated methods on the four datasets with and without our process. The results on SUN~\cite{sun} are again mostly consistent with what has been observed on CUB~\cite{cub} and AwA2~\cite{xian2017arxiv}, while results on aPY~\cite{apy} differ due to the nature of the dataset. In particular, due to the difficulty in selecting good hyperparameters on this dataset, the regularization parameter selected for the GZSL task is not always relevant and does not lead to an increase in the final score.

Table~\ref{table:sup_10crop} is an extended version of Table~3 in the main text. It reports results for the ZSL and the GZSL tasks when visual features are improved with 10-crop. The AUSUC metric~\cite{chao2016} has been added for GZSL; the accuracies used for computing the Seen-Unseen Curve are the per sample accuracies as is usually the case. 
Visual features obtained with 10-crop almost always lead to better results on CUB~\cite{cub}, AwA2~\cite{xian2017arxiv} and SUN~\cite{sun}, while again no clear pattern is visible for aPY~\cite{apy}. 
It is worth noting that although the harmonic mean score and the AUSUC score are visibly correlated, a better AUSUC does not necessarily imply a better harmonic mean. For example, among the bilinear compatibility models ALE~\cite{ale}, DeViSE~\cite{devise} and SJE~\cite{sje} on the AwA2~\cite{xian2017arxiv} dataset, DeViSE has the highest AUSUC score with 0.538 (compared to respectively 0.512 and 0.529 for ALE and SJE) but the lowest harmonic mean score with 55.0 (per class accuracy, compared to respectively 56.9 and 59.4 for ALE and SJE).

\begin{table}
\centering
\caption{Previously reported and independently reproduced ZSL per-class accuracy. S.S.: Standard Split, P.S.: Proposed Split~\cite{xian2017cvpr}. Averaged over 5 runs.}
\label{table:sup_zsl}
\begin{tabular}{|l|c c|c c|}
\hline
\multirow{2}{*}{\textbf{Method}} & \multicolumn{2}{|c|}{\textbf{Reported in}~\cite{xian2017cvpr}} & \multicolumn{2}{|c|}{\textbf{Reproduced}} \\
& \textbf{S.S.} & \textbf{P.S.} & \textbf{S.S.} & \textbf{P.S.} \\
\hline\hline
\multicolumn{5}{|c|}{\textbf{CUB}~\cite{cub}} \\
\hline
Linear\textsubscript{$\mathcal{V} \rightarrow \mathcal{S}$} & \emph{n/a} & \emph{n/a} & 41.0 $\pm$ 0.0 & 41.8 $\pm$ 0.0 \\
Linear\textsubscript{$\mathcal{S} \rightarrow \mathcal{V}$} & \emph{n/a} & \emph{n/a} & 56.0 $\pm$ 0.0 & 53.5 $\pm$ 0.0 \\
ALE~\cite{ale} & 53.2 & 54.9 & 54.8 $\pm$ 0.8 & 54.0 $\pm$ 1.2 \\
DeViSE~\cite{devise} & 53.2 & 52.0 & 52.5 $\pm$ 0.9 & 52.6 $\pm$ 1.3 \\
SJE~\cite{sje} & \textbf{55.3} & 53.9 & 53.8 $\pm$ 2.3 & 49.2 $\pm$ 1.4 \\
ESZSL~\cite{eszsl} & 55.1 & 53.9 & 34.9 $\pm$ 0.0 & 34.9 $\pm$ 0.0 \\
Sync~\cite{sync} & 54.1 & \textbf{55.6} & \textbf{56.4} $\pm$ 0.9 & \textbf{54.8} $\pm$ 0.6 \\
SAE~\cite{sae} & 33.4 & 33.3 & 56.2 $\pm$ 0.0 & 53.3 $\pm$ 0.0 \\
\hline\hline
\multicolumn{5}{|c|}{\textbf{AwA2}~\cite{xian2017arxiv}} \\
\hline
Linear\textsubscript{$\mathcal{V} \rightarrow \mathcal{S}$} & \emph{n/a} & \emph{n/a} & 68.2 $\pm$ 0.0 & 49.7 $\pm$ 0.0 \\
Linear\textsubscript{$\mathcal{S} \rightarrow \mathcal{V}$} & \emph{n/a} & \emph{n/a} & \textbf{85.5} $\pm$ 0.0 & \textbf{68.9} $\pm$ 0.0 \\
ALE~\cite{ale} & 80.3 & \textbf{62.5} & 80.3 $\pm$ 2.2 & 62.9 $\pm$ 2.3 \\
DeViSE~\cite{devise} & 68.6 & 59.7 & 76.6 $\pm$ 1.6 & 62.1 $\pm$ 1.6 \\
SJE~\cite{sje} & 69.5 & 61.9 & 80.4 $\pm$ 2.9 & 62.2 $\pm$ 1.2 \\
ESZSL~\cite{eszsl} & 75.6 & 58.6 & 70.5 $\pm$ 0.0 & 50.8 $\pm$ 0.0 \\
Sync~\cite{sync} & 71.2 & 46.6 & 65.6 $\pm$ 0.8 & 58.1 $\pm$ 0.8 \\
SAE~\cite{sae} & \textbf{80.7} & 54.1 & 81.1 $\pm$ 0.0 & 62.8 $\pm$ 0.0 \\
\hline\hline
\multicolumn{5}{|c|}{\textbf{SUN}~\cite{sun}} \\
\hline
Linear\textsubscript{$\mathcal{V} \rightarrow \mathcal{S}$} & \emph{n/a} & \emph{n/a} & 50.8 $\pm$ 0.0 & 46.7 $\pm$ 0.0 \\
Linear\textsubscript{$\mathcal{S} \rightarrow \mathcal{V}$} & \emph{n/a} & \emph{n/a} & 62.8 $\pm$ 0.0 & \textbf{61.5} $\pm$ 0.0 \\
ALE~\cite{ale} & 59.1 & \textbf{58.1} & 63.5 $\pm$ 0.5 & 59.4 $\pm$ 0.3 \\
DeViSE~\cite{devise} & 57.5 & 56.5 & 61.6 $\pm$ 0.3 & 58.4 $\pm$ 0.3 \\
SJE~\cite{sje} & 57.1 & 53.7 & 59.9 $\pm$ 0.6 & 55.3 $\pm$ 1.1 \\
ESZSL~\cite{eszsl} & 57.3 & 54.5 & 21.0 $\pm$ 0.0 & 16.0 $\pm$ 0.0 \\
Sync~\cite{sync} & \textbf{59.1} & 56.3 & 58.5 $\pm$ 0.5 & 56.6 $\pm$ 1.8 \\
SAE~\cite{sae} & 42.4 & 40.3 & \textbf{63.7} $\pm$ 0.0 & 61.0 $\pm$ 0.0 \\
\hline\hline
\multicolumn{5}{|c|}{\textbf{aPY}~\cite{apy}} \\
\hline
Linear\textsubscript{$\mathcal{V} \rightarrow \mathcal{S}$} & \emph{n/a} & \emph{n/a} & 30.0 $\pm$ 0.0 & 31.6 $\pm$ 0.0 \\
Linear\textsubscript{$\mathcal{S} \rightarrow \mathcal{V}$} & \emph{n/a} & \emph{n/a} & \textbf{42.2} $\pm$ 0.0 & \textbf{40.0} $\pm$ 0.0 \\
ALE~\cite{ale} & 30.9 & 39.7 & 16.0 $\pm$ 5.7 & 30.4 $\pm$ 0.9 \\
DeViSE~\cite{devise} & 35.4 & \textbf{39.8} & 30.3 $\pm$ 3.2 & 29.0 $\pm$ 1.1 \\
SJE~\cite{sje} & 32.0 & 32.9 & 18.9 $\pm$ 2.7 & 33.5 $\pm$ 1.9 \\
ESZSL~\cite{eszsl} & 34.4 & 38.3 & 31.6 $\pm$ 0.0 & 16.3 $\pm$ 0.0 \\
Sync~\cite{sync} & \textbf{39.7} & 23.9 & 34.1 $\pm$ 0.5 & 35.5 $\pm$ 0.1 \\
SAE~\cite{sae} & 8.3 & 8.3 & 31.7 $\pm$ 0.0 & 29.0 $\pm$ 0.0 \\
\hline
\end{tabular}
\end{table}

\begin{table}
\centering
\caption{GZSL score (harmonic mean of $A_{\mathcal{U} \rightarrow \mathcal{C}}$ and $A_{\mathcal{S} \rightarrow \mathcal{C}}$, per class accuracy) with and without calibration and GZSL regularization. On Proposed Split~\cite{xian2017cvpr}, averaged over 5 runs.}
\label{table:sup_gzsl}
\begin{tabular}{|l|c|c c c|}
\hline
\textbf{Method} & \textbf{Reported in} \cite{xian2017cvpr} & \multicolumn{3}{|c|}{\textbf{Ours}} \\
 \hline
 with \textbf{calibration}             & - & - & \cmark & \cmark \\
 with $\boldsymbol{\lambda}^{*}_{\mathbf{GZSL}}$ & - & - & - & \cmark \\
\hline
\multicolumn{5}{|c|}{\textbf{CUB}~\cite{cub}} \\
\hline
Linear\textsubscript{$\mathcal{V} \rightarrow \mathcal{S}$} & \emph{n/a} & 18.2 $\pm$ 0.0 & 34.3 $\pm$ 0.0 & 35.5 $\pm$ 0.0 \\
Linear\textsubscript{$\mathcal{S} \rightarrow \mathcal{V}$} & \emph{n/a} & 32.5 $\pm$ 0.0 & 41.9 $\pm$ 0.0 & 43.5 $\pm$ 0.0 \\
ALE~\cite{ale} & 34.4 & \textbf{35.6} $\pm$ 1.3 & 45.1 $\pm$ 1.3 & 46.2 $\pm$ 1.0 \\
DeViSE~\cite{devise} & 32.8 & 35.1 $\pm$ 0.6 & 43.6 $\pm$ 1.1 & 43.4 $\pm$ 0.7 \\
SJE~\cite{sje} & 33.6 & 29.7 $\pm$ 1.7 & 41.2 $\pm$ 0.6 & 44.2 $\pm$ 1.0 \\
ESZSL~\cite{eszsl} & 21.0 & 17.9 $\pm$ 0.0 & 33.7 $\pm$ 0.0 & 33.9 $\pm$ 0.0 \\
Sync~\cite{sync} & 19.8 & 33.2 $\pm$ 1.0 & \textbf{46.2} $\pm$ 0.8 & \textbf{47.6} $\pm$ 0.2 \\
SAE~\cite{sae} & 13.6 & 25.7 $\pm$ 0.0 & 43.1 $\pm$ 0.0 & 43.1 $\pm$ 0.0 \\
\hline
Average (CUB) & 25.9 & 28.5 & 41.1  & \textbf{42.2} \\
\hline\hline
\multicolumn{5}{|c|}{\textbf{AwA2}~\cite{xian2017arxiv}} \\
\hline
Linear\textsubscript{$\mathcal{V} \rightarrow \mathcal{S}$} & \emph{n/a} & 8.3 $\pm$ 0.0 & 47.3 $\pm$ 0.0 & 48.1 $\pm$ 0.0 \\
Linear\textsubscript{$\mathcal{S} \rightarrow \mathcal{V}$} & \emph{n/a} & \textbf{44.3} $\pm$ 0.0 & \textbf{62.7} $\pm$ 0.0 & \textbf{64.0} $\pm$ 0.0 \\
ALE~\cite{ale} & 23.9 & 26.9 $\pm$ 1.5 & 55.8 $\pm$ 1.7 & 55.8 $\pm$ 1.7 \\
DeViSE~\cite{devise} & 27.8 & 17.4 $\pm$ 4.1 & 54.6 $\pm$ 1.2 & 54.6 $\pm$ 1.2 \\
SJE~\cite{sje} & 14.4 & 28.9 $\pm$ 2.3 & 58.2 $\pm$ 0.9 & 59.0 $\pm$ 1.9 \\
ESZSL~\cite{eszsl} & 11.0 & 39.9 $\pm$ 0.0 & 53.6 $\pm$ 0.0 & 53.7 $\pm$ 0.0 \\
Sync~\cite{sync} & 18.0 & 30.6 $\pm$ 0.3 & 61.0 $\pm$ 0.5 & 61.0 $\pm$ 0.5 \\
SAE~\cite{sae} & 2.2 & 29.5 $\pm$ 0.0 & 60.2 $\pm$ 0.0 & 60.2 $\pm$ 0.0 \\
\hline
Average (AwA2) & 16.2 & 28.2 & 56.7 & \textbf{57.1}\\
\hline\hline
\multicolumn{5}{|c|}{\textbf{SUN}~\cite{sun}} \\
\hline
Linear\textsubscript{$\mathcal{V} \rightarrow \mathcal{S}$} & \emph{n/a} & 16.3 $\pm$ 0.0 & 23.3 $\pm$ 0.0 & 25.2 $\pm$ 0.0 \\
Linear\textsubscript{$\mathcal{S} \rightarrow \mathcal{V}$} & \emph{n/a} & 24.4 $\pm$ 0.0 & 34.2 $\pm$ 0.0 & 34.8 $\pm$ 0.0 \\
ALE~\cite{ale} & 26.3 & \textbf{26.9} $\pm$ 0.3 & 33.6 $\pm$ 0.2 & 33.6 $\pm$ 0.2 \\
DeViSE~\cite{devise} & 20.9 & 25.2 $\pm$ 0.2 & 31.4 $\pm$ 0.4 & 31.4 $\pm$ 0.4 \\
SJE~\cite{sje} & 19.8 & 25.5 $\pm$ 0.6 & \textbf{34.9} $\pm$ 0.4 & \textbf{35.3} $\pm$ 0.4 \\
ESZSL~\cite{eszsl} & 15.8 & 6.7 $\pm$ 0.0 & 11.1 $\pm$ 0.0 & 11.1 $\pm$ 0.0 \\
Sync~\cite{sync} & 13.4 & 20.6 $\pm$ 0.8 & 27.9 $\pm$ 1.0 & 27.9 $\pm$ 1.0 \\
SAE~\cite{sae} & 11.8 & 25.1 $\pm$ 0.0 & 34.0 $\pm$ 0.0 & 34.7 $\pm$ 0.0 \\
\hline
Average (SUN) & 18.0 & 21.3 & 28.8 & \textbf{29.2}\\
\hline\hline
\multicolumn{5}{|c|}{\textbf{aPY}~\cite{apy}} \\
\hline
Linear\textsubscript{$\mathcal{V} \rightarrow \mathcal{S}$} & \emph{n/a} & 4.0 $\pm$ 0.0 & 32.7 $\pm$ 0.0 & 32.7 $\pm$ 0.0 \\
Linear\textsubscript{$\mathcal{S} \rightarrow \mathcal{V}$} & \emph{n/a} & \textbf{21.5} $\pm$ 0.0 & \textbf{39.1} $\pm$ 0.0 & \textbf{39.2} $\pm$ 0.0 \\
ALE~\cite{ale} & 8.7 & 19.4 $\pm$ 0.5 & 30.1 $\pm$ 0.9 & 29.6 $\pm$ 1.1 \\
DeViSE~\cite{devise} & 9.2 & 20.7 $\pm$ 0.7 & 28.0 $\pm$ 1.0 & 27.4 $\pm$ 1.1 \\
SJE~\cite{sje} & 6.9 & 16.1 $\pm$ 1.3 & 36.3 $\pm$ 1.2 & 36.3 $\pm$ 1.2 \\
ESZSL~\cite{eszsl} & 4.6 & 13.3 $\pm$ 0.0 & 20.6 $\pm$ 0.0 & 20.4 $\pm$ 0.0 \\
Sync~\cite{sync} & \emph{\textbf{13.3}} & 13.8 $\pm$ 0.1 & 36.4 $\pm$ 0.1 & 36.4 $\pm$ 0.1 \\
SAE~\cite{sae} & 0.9 & 16.7 $\pm$ 0.0 & 33.8 $\pm$ 0.0 & 32.1 $\pm$ 0.0 \\
\hline
Average (aPY) & 7.3 & 15.7 & \textbf{32.1} & 31.8 \\
\hline
\end{tabular}
\end{table}

\begin{table}
\centering
\caption{ZSL and GZSL scores with 10-crop features, evaluated with per class (p.c.) and per sample (p.s.) accuracies. With calibration and $\lambda^{*}_{GZSL}$. On P.S.~\cite{xian2017cvpr}, averaged over 5 runs.}
\label{table:sup_10crop}
\begin{tabular}{|l|c c|c c c|}
\hline
\multirow{2}{*}{\textbf{Method}} & \multicolumn{2}{|c|}{\textbf{ZSL}} & \multicolumn{3}{|c|}{\textbf{GZSL}} \\
& \textbf{Acc. p.c.} & \textbf{Acc. p.s.} & \textbf{H. p.c.} & \textbf{H. p.s.} & \textbf{AUSUC}\\
\hline\hline
\multicolumn{6}{|c|}{\textbf{CUB}~\cite{cub}} \\
\hline
Linear\textsubscript{$\mathcal{V} \rightarrow \mathcal{S}$} & 45.6 $\pm$ 0.0 & 45.6 $\pm$ 0.0 & 39.8 $\pm$ 0.0 & 39.8 $\pm$ 0.0 & 0.246 $\pm$ 0.000 \\
Linear\textsubscript{$\mathcal{S} \rightarrow \mathcal{V}$} & 57.1 $\pm$ 0.0 & 57.2 $\pm$ 0.0 & 47.7 $\pm$ 0.0 & 48.0 $\pm$ 0.0 & 0.315 $\pm$ 0.000 \\
ALE~\cite{ale} & 57.4 $\pm$ 0.3 & 57.5 $\pm$ 0.3 & \textbf{49.2} $\pm$ 0.6 & \textbf{49.3} $\pm$ 0.6 & \textbf{0.347} $\pm$ 0.005 \\
DeViSE~\cite{devise} & 52.9 $\pm$ 1.0 & 52.9 $\pm$ 1.0 & 42.4 $\pm$ 0.8 & 42.5 $\pm$ 0.7 & 0.270 $\pm$ 0.007 \\
SJE~\cite{sje} & 51.9 $\pm$ 1.7 & 52.1 $\pm$ 1.7 & 46.7 $\pm$ 0.9 & 46.9 $\pm$ 1.0 & 0.308 $\pm$ 0.009 \\
ESZSL~\cite{eszsl} & 39.0 $\pm$ 0.0 & 38.8 $\pm$ 0.0 & 38.7 $\pm$ 0.0 & 38.6 $\pm$ 0.0 & 0.224 $\pm$ 0.000 \\
Sync~\cite{sync} & \textbf{57.5} $\pm$ 1.8 & \textbf{57.6} $\pm$ 1.8 & 48.9 $\pm$ 1.2 & 49.1 $\pm$ 1.2 & 0.337 $\pm$ 0.018 \\
SAE~\cite{sae} & 56.1 $\pm$ 0.0 & 56.2 $\pm$ 0.0 & 46.3 $\pm$ 0.0 & 46.6 $\pm$ 0.0 & 0.307 $\pm$ 0.000 \\
\hline\hline
\multicolumn{6}{|c|}{\textbf{AwA2}~\cite{xian2017arxiv}} \\
\hline
Linear\textsubscript{$\mathcal{V} \rightarrow \mathcal{S}$} & 51.0 $\pm$ 0.0 & 43.6 $\pm$ 0.0 & 49.0 $\pm$ 0.0 & 45.5 $\pm$ 0.0 & 0.357 $\pm$ 0.000 \\
Linear\textsubscript{$\mathcal{S} \rightarrow \mathcal{V}$} & \textbf{70.4} $\pm$ 0.0 & \textbf{69.3} $\pm$ 0.0 & \textbf{65.1} $\pm$ 0.0 & \textbf{68.7} $\pm$ 0.0 & \textbf{0.598} $\pm$ 0.000 \\
ALE~\cite{ale} & 63.0 $\pm$ 1.8 & 61.1 $\pm$ 1.7 & 56.9 $\pm$ 1.8 & 55.5 $\pm$ 2.2 & 0.521 $\pm$ 0.014 \\
DeViSE~\cite{devise} & 63.1 $\pm$ 1.8 & 62.2 $\pm$ 0.6 & 55.0 $\pm$ 1.7 & 50.6 $\pm$ 0.8 & 0.538 $\pm$ 0.005 \\
SJE~\cite{sje} & 63.8 $\pm$ 2.0 & 61.6 $\pm$ 2.8 & 59.4 $\pm$ 1.2 & 57.6 $\pm$ 1.8 & 0.529 $\pm$ 0.022 \\
ESZSL~\cite{eszsl} & 52.6 $\pm$ 0.0 & 51.9 $\pm$ 0.0 & 54.4 $\pm$ 0.0 & 57.9 $\pm$ 0.0 & 0.434 $\pm$ 0.000 \\
Sync~\cite{sync} & 59.3 $\pm$ 0.2 & 56.1 $\pm$ 0.4 & 62.6 $\pm$ 0.1 & 63.2 $\pm$ 0.2 & 0.511 $\pm$ 0.003 \\
SAE~\cite{sae} & 63.5 $\pm$ 0.0 & 65.4 $\pm$ 0.0 & 62.3 $\pm$ 0.0 & 63.6 $\pm$ 0.0 & 0.585 $\pm$ 0.000 \\
\hline\hline
\multicolumn{6}{|c|}{\textbf{SUN}~\cite{sun}} \\
\hline
Linear\textsubscript{$\mathcal{V} \rightarrow \mathcal{S}$} & 48.6 $\pm$ 0.0 & 48.6 $\pm$ 0.0 & 26.7 $\pm$ 0.0 & 25.8 $\pm$ 0.0 & 0.107 $\pm$ 0.000 \\
Linear\textsubscript{$\mathcal{S} \rightarrow \mathcal{V}$} & 62.0 $\pm$ 0.0 & 62.0 $\pm$ 0.0 & 36.3 $\pm$ 0.0 & 35.4 $\pm$ 0.0 & 0.184 $\pm$ 0.000 \\
ALE~\cite{ale} & 62.2 $\pm$ 0.2 & 62.2 $\pm$ 0.2 & 34.9 $\pm$ 0.2 & 34.4 $\pm$ 0.3 & 0.183 $\pm$ 0.002 \\
DeViSE~\cite{devise} & 61.2 $\pm$ 0.3 & 61.2 $\pm$ 0.3 & 32.5 $\pm$ 0.4 & 32.1 $\pm$ 0.4 & 0.167 $\pm$ 0.003 \\
SJE~\cite{sje} & 58.2 $\pm$ 0.8 & 58.4 $\pm$ 0.8 & \textbf{36.8} $\pm$ 0.6 & \textbf{36.0} $\pm$ 0.4 & \textbf{0.194} $\pm$ 0.002 \\
ESZSL~\cite{eszsl} & 17.2 $\pm$ 0.0 & 17.2 $\pm$ 0.0 & 11.8 $\pm$ 0.0 & 11.6 $\pm$ 0.0 & 0.169 $\pm$ 0.000 \\
Sync~\cite{sync} & 56.6 $\pm$ 1.8 & 56.6 $\pm$ 1.8 & 27.9 $\pm$ 1.0 & 27.2 $\pm$ 1.1 & 0.123 $\pm$ 0.008 \\
SAE~\cite{sae} & \textbf{62.7} $\pm$ 0.0 & \textbf{62.7} $\pm$ 0.0 & 35.9 $\pm$ 0.0 & 35.1 $\pm$ 0.0 & 0.185 $\pm$ 0.000 \\
\hline\hline
\multicolumn{6}{|c|}{\textbf{aPY}~\cite{apy}} \\
\hline
Linear\textsubscript{$\mathcal{V} \rightarrow \mathcal{S}$} & 28.6 $\pm$ 0.0 & 12.5 $\pm$ 0.0 & 33.3 $\pm$ 0.0 & 16.2 $\pm$ 0.0 & 0.081 $\pm$ 0.000 \\
Linear\textsubscript{$\mathcal{S} \rightarrow \mathcal{V}$} & \textbf{40.6} $\pm$ 0.0 & \textbf{23.9} $\pm$ 0.0 & \textbf{38.4} $\pm$ 0.0 & 21.0 $\pm$ 0.0 & 0.098 $\pm$ 0.000 \\
ALE~\cite{ale} & 31.3 $\pm$ 1.0 & 16.8 $\pm$ 2.7 & 31.5 $\pm$ 0.9 & 21.1 $\pm$ 1.0 & 0.101 $\pm$ 0.005 \\
DeViSE~\cite{devise} & 30.9 $\pm$ 0.9 & 16.4 $\pm$ 1.6 & 28.7 $\pm$ 1.2 & 19.8 $\pm$ 0.8 & 0.096 $\pm$ 0.005 \\
SJE~\cite{sje} & 34.1 $\pm$ 0.8 & 16.4 $\pm$ 1.6 & 36.9 $\pm$ 0.7 & \textbf{23.0} $\pm$ 1.2 & \textbf{0.110} $\pm$ 0.009 \\
ESZSL~\cite{eszsl} & 12.2 $\pm$ 0.0 & 7.2 $\pm$ 0.0 & 17.1 $\pm$ 0.0 & 11.6 $\pm$ 0.0 & 0.077 $\pm$ 0.000 \\
Sync~\cite{sync} & 34.9 $\pm$ 0.2 & 16.8 $\pm$ 0.2 & 37.4 $\pm$ 0.1 & 19.4 $\pm$ 0.1 & 0.102 $\pm$ 0.001 \\
SAE~\cite{sae} & 29.5 $\pm$ 0.0 & 11.1 $\pm$ 0.0 & 32.7 $\pm$ 0.0 & 13.0 $\pm$ 0.0 & 0.058 $\pm$ 0.000 \\
\hline
\end{tabular}
\end{table}

\end{document}